\documentclass{article}

\usepackage{arxiv}

\usepackage[utf8]{inputenc} % allow utf-8 input
\usepackage[T1]{fontenc}    % use 8-bit T1 fonts
\usepackage{hyperref}       % hyperlinks
\usepackage{url}            % simple URL typesetting
\usepackage{booktabs}       % professional-quality tables
\usepackage{amsfonts}       % blackboard math symbols
\usepackage{nicefrac}       % compact symbols for 1/2, etc.
\usepackage{microtype}      % microtypography
\usepackage{lipsum}		% Can be removed after putting your text content
\usepackage{graphicx}
\usepackage[numbers]{natbib}
\usepackage{doi}
\usepackage{times}
\usepackage{multirow}
\usepackage{mathtools}
\newcommand{\indep}{\perp\!\!\!\!\perp}
\newcommand\inner[2]{\langle #1, #2 \rangle}
\usepackage{amsmath}
\usepackage{amssymb}
\usepackage{natbib}
\usepackage{graphicx}
\usepackage{url}
\usepackage{algorithm2e}
\usepackage{subfigure}

\title{Causal Effect Estimation Using Random Hyperplane Tessellations}

\author{
	Abhishek Dalvi 
     \And
	Neil Ashtekar 
	\And
	Vasant Honavar 
}

\renewcommand{\shorttitle}{\textit{Causal Effect Estimation Using Random Hyperplane Tessellations}}

\hypersetup{
pdftitle={Causal Effect Estimation Using Random Hyperplane Tessellations(preprint)},
pdfsubject={ML},
pdfauthor={Abhishek Dalvi},
pdfkeywords={Causality, Matching},
}

\begin{document}
\maketitle

\begin{abstract}
	Matching is one of the simplest approaches for estimating causal effects from observational data. Matching techniques compare the observed outcomes across pairs of individuals with similar covariate values but different treatment statuses in order to estimate causal effects. However, traditional matching techniques are unreliable given high-dimensional covariates due to the infamous curse of dimensionality. To overcome this challenge, we propose a simple, fast, yet highly effective approach to matching using Random Hyperplane Tessellations (RHPT). First, we prove that the RHPT representation is an approximate balancing score -- thus maintaining the strong ignorability assumption -- and provide empirical evidence for this claim. Second, we report results of extensive experiments showing that matching using RHPT outperforms traditional matching techniques and is competitive with state-of-the-art deep learning methods for causal effect estimation. In addition, RHPT avoids the need for computationally expensive training of deep neural networks.
\end{abstract}

% keywords can be removed

\section{Introduction}

Randomized experiments, whenever feasible, provide the most reliable estimates of causal effects. In a randomized experiment, each individual is randomly assigned to either the treatment or control group, thereby ensuring covariate balance: the two groups will be only {\em randomly different} from each other with respect to all covariates, both observed and unobserved. In this case, it is straightforward to show that the average treatment effect $\text{ATE} = \mathbb {E}[Y|T=1]- \mathbb {E}[Y |T=0]$, where $Y$ is the observed outcome and $T$ is the treatment status. In other words, the ATE is simply the difference in average \textit{observed} outcomes across the treatment and control groups. In the case of a randomized experiment, association is causation.  

However, randomized experiments are not always feasible due to their high costs, ethical concerns, or other fundamental limitations. As such, there is high interest in methods capable of reliably estimating causal effects from observational data. The following assumptions suffice for estimating causal effects from observational data: {\em Ignorability}, {\em Positivity}, {\em Non-interference} and {\em Consistency} \citep{balacningrsenbaum, no_interference,SUTVA_cite}. Under these conditions, it can be shown that: \[ \text{ATE}= \mathbb {E}_{\bf X} [\mathbb {E}[Y|T=1, {\bf X}={\bf x}]- \mathbb {E}[Y |T=0,{\bf X}={\bf x}]] \]  

In observational data the treatment assignment is not randomized, it is possible that some of the covariates influence both the treatment assignment and the outcome; thus, introducing confounding bias while estimating the causal effect. In the presence of confounding, causation is {\em not} association. One way to eliminate confounding bias under the standard assumptions is to \textit{simulate} a randomized experiment such that the treatment and control groups have similar covariate distributions.

This can be accomplished through matching individuals with similar covariates across the treated and control groups, thereby reducing bias in treatment assignment \citep{Hernan,Stuart2010MatchingMF}. Note that matching on the observed covariates also matches on the unobserved covariates, in so much as they are correlated with those that are observed\footnote{Bias introduced by unobserved covariates that are uncorrelated with the observed covariates is assessed using sensitivity analyses \citep{Hernan}.}.

Comparing the observed outcome for an individual with covariates $x_i$ with treatment status $T_i = 1$ against the individual they are matched with, $m(x_i)$ with treatment status $T_{m(x_i)} = 0$ yields an estimate for the individual treatment effect $\text{ITE}(x_i)$. Matching techniques have been extensively studied in the literature on causal inference -- see \citet{Stuart2010MatchingMF} for a review.

Because perfect matches can seldom be found, matching in practice entails nearest neighbor search. However, matching techniques typically perform poorly when the number of covariates is large, since all points tend to appear dissimilar in high-dimensional spaces. This phenomenon is known as the curse of dimensionality \citep{Beyer1999WhenI}. Empirical results from \citet{Beyer1999WhenI} and \citet{Aggarwal2001OnTS} suggest that distance metrics become unreliable given as few as 10-20 dimensions.

A simple, popular workaround is to match individuals based on their propensity scores -- an individual's probability of being treated -- and to rely on the theoretical result that once conditioned on the propensity scores, treatment becomes independent of the covariates, thereby satisfying the identifiability conditions. In practice, matching on propensity scores seldom achieves its intended objective \citep{king2019propensity}. 
Recently there has been increased interest in  deep neural networks that learn low-dimensional, ``treatment-agnostic'' representations of high-dimensional data using non-linear mappings \citep{CFRnet,TARNET, CEVAE, koch_sainburg_geraldo_jiang_sun_foster_2021}. While deep learning approaches offer strong performance, they require large amounts of training data, incur substantial computational overhead for training, and require expensive hyper-parameter tuning. 

In this work, we explore Random Hyperplane Tessellations (RHPT) an alternative approach which overcomes the limitations of both existing matching techniques as well as deep neural networks for causal effect estimation.  RHPT involves mapping covariates to a high-dimensional, binary space based on the relationship between covariate values and randomly chosen hyperplanes \citep{Vershyninhyperplane_tessla, Dirksensharp_randomHyperPlane}. The RHPT representation is attractive for causal effect estimation because it offers a compromise between information-poor (and potentially inaccurate) predicted propensity scores and information-rich (but potentially noisy) matching on covariates. RHPT can be viewed as a denoising, information-reduction step performed prior to matching. 

The key contributions of this paper are as follows:

\begin{itemize}
\item We present a novel application of Random Hyperplane Tessellations for estimating causal effects from observational data and explain via theory why it works.

\item We show that our method preserves strong ignorability of treatment assignments as our RHPT representations are approximate balancing scores.

\item We present experiments comparing treatment effect estimates obtained using our proposed method to estimates obtained using traditional matching approaches as well as state-of-the-art deep learning techniques. The results of these experiments on several standard benchmark data sets show that our proposed method outperforms traditional matching approaches and is competitive with the state-of-the-art deep neural networks in terms of accuracy while avoiding the substantial computational cost of training deep neural networks.
\end{itemize}

\section{Preliminaries}\label{Preliminaries}

\subsection{Problem Definition}

We are given data on $n$ individuals each represented as a tuple $(\mathbf{x}_i, T_i, Y_i)$. $\mathbf{x}_i$ denotes the (pre-treatment) covariates, $T_i$ denotes treatment status ($T_i = 1$ indicating treatment and $T_i = 0$ indicating control), and $Y_i$ denotes the observed outcome. Formally, $\mathbf{x}_i \in \mathcal{X}$ with $ \mathcal{X} \subseteq \mathbb{R}^{d}$, where $d$ is the dimensionality of the covariates, $T_i \in \{0,1\}$, and $Y_i \in \mathcal{Y}$ with $\mathcal{Y} \subseteq \mathbb{R}$. Our goal is to estimate the $\text{ATE}$ across all individuals and the $\text{ITE}_i$ for each individuals $i$ under the strong ignorability, consistency, and non-interference assumptions.

\subsection{Locality Sensitive hashing}

Locality-Sensitive Hashing (LSH) \citep{IndykLSH} was one of the first works related to hyperplane tessellations and is a widely used approach for approximately finding nearest neighbors in high-dimensional spaces. A locality sensitive hashing scheme is a distribution on a family ${\mathcal H}$ of hash functions\footnote{A \textit{hash function} maps input samples to fixed-sized output hash codes. We consider Locality-sensitive hashing (LSH); a fuzzy hashing technique that hashes similar input items into the same to similar hash codes.} $h$ operating on a collection of objects, such that for two objects ${\mathbf x_i}$ and ${\mathbf x_j}$, $\Pr_{h \in {\mathcal H}}[h({\mathbf x_i}) = h({\mathbf x_j})] = sim({\mathbf x_i}, {\mathbf x_j}) \in [0,1]$, where $ sim({\mathbf x_i}, {\mathbf x_j})$ is a similarity function. Such a scheme leads to a compact representation of objects such that the similarity of objects can be estimated from their compact sketches.

Given covariates $\mathbf{x}_i \in \mathbb{R}^{d}$, SimHash \citep{cahrikar} generates a random hyperplane $\mathbf{a}^{\perp} \in \mathbb{R}^{d}$ using a standard Gaussian random vector $\mathbf{a} \sim \mathcal{N}(0,I_{d})$.  
Using random vector $\mathbf{a}$ and function $\text{sign}(\cdot) : \mathbb{R} \rightarrow \{0,1\}$, we generate a hash value for $\mathbf{x}_i$:

$$ h_{\mathbf{a}}(\mathbf{x}_i) = \text{sign} (\mathbf{a} \cdot \mathbf{x}_i)$$

\noindent with $\Pr[h_{\mathbf{r}}(\mathbf{x}_i) = h_{\mathbf{r}}(\mathbf{x}_j)] =  1 - \frac{\theta_{ij}}{\pi}$ where $\theta_{ij}$ is the angle between $\mathbf{x}_i$ and $\mathbf{x}_j$. A binary sketch can be obtained by performing hashing $\beta$ times. Let $\mathbf{A}: \mathbb{R}^{d} \rightarrow \mathbb{R}^{\beta}$ be a matrix with each row $\mathbf{a}_{1}^{T}, \cdots \mathbf{a}_{\beta}^{T}$ drawn from  $\mathcal{N}(0,I_{d})$. For conciseness, denote $\mathbf{A} \sim \mathcal{N}^{\beta}(0,I_{d})$ and apply the sign function elementwise, i.e. $\text{sign}(\cdot): \mathbb{R}^{\beta} \rightarrow \{0,1\}^{\beta}$. The binary sketch obtained using SimHash is $ h_{\mathbf{A}}(\mathbf{x}_i) = \text{sign} (\mathbf{A} \mathbf{x}_i)$.

\subsection{Random Hyper-Plane Tessellations (RHPT)}

\begin{figure}[t]
\includegraphics[width=8cm]{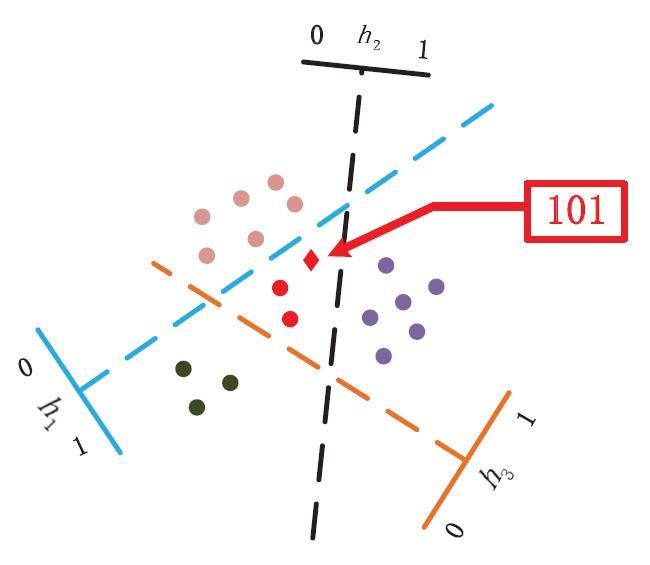}
\centering
\caption{An example of a Random Hyperplane
Tessellation (RHPT) for a set of points in a two-dimensional space from \cite{RHPT_figure_cite}. Three hyperplanes -- indicated by dashed lines -- are used to construct a three-dimensional binary hash code for each point. The color of each point represents its binary hash code: red represents $101$, black represents $100$, etc.}
\end{figure}

\citet{Vershyninhyperplane_tessla} showed that Simhash results in a representation where angular distances are preserved with  error $\leq \delta$; given that the points are subset of a unit Euclidean sphere $S^{d-1}$. Formally, we are given $\mathcal{X} \subseteq S^{d-1}$, positive constants $c_{1}$ and $c_{2}$, $\ell_{*}(\mathcal{X}) = \mathbb{E} \; \underset{\mathbf{x} \in \mathcal{X}}{\sup}|\inner{\mathbf{a}}{\mathbf{x}}|$, and $\delta > 0$. If the following holds:

\begin{equation*}
    \beta \geq c_{1}\frac{\ell^{2}_{*}(\mathcal{X})}{\delta^{6}}
\end{equation*}

\noindent then with a probability of at least $1 - 2\exp(-c_{2}\beta \delta^{2})$, for every $\mathbf{x}_{i}, \mathbf{x}_{j} \in \mathcal{X}$:

\begin{equation} \label{anglehash}
    |\frac{1}{\beta}d_{H}[\text{sign}(\mathbf{A}\mathbf{x}_i),\text{sign}(\mathbf{A}\mathbf{x}_j )]- d_{S^{d-1}}[\mathbf{x}_i,\mathbf{x}_j]| \leq \delta
\end{equation}

\noindent where $d_{H}$ is the Hamming distance and $d_{S^{d-1}}[\mathbf{x}_i,\mathbf{x}_j]| = \frac{1}{\pi}\arccos{\frac{\inner{\mathbf{x}_i}{\mathbf{x}_j}}{||\mathbf{x}_i||_{2} ||\mathbf{x}_j||_{2}}}$ is the angular distance. This is consistent with the findings in \cite{cahrikar, simhash_scaling} which state that:

$$\cos(\theta_{ij}) \approx \cos(\frac{\pi}{\beta}d_{H}[\text{sign}(\mathbf{A}\mathbf{x}_i),\text{sign}(\mathbf{A}\mathbf{x}_j )])$$

\noindent To summarize, $\delta \rightarrow 0$ with probability $\rightarrow 1$ when $\beta \rightarrow \infty$. In other words, distances are approximately  preserved as the dimensionality of the embedding is increased.

\citet{Dirksensharp_randomHyperPlane} extend the results from \citet{Vershyninhyperplane_tessla} beyond the unit Euclidean sphere $S^{d-1}$ to an arbitrary set $\mathbb{R}^{d}$ by introducing shifts on the hyperplane represented by $\gamma$. Let $\mathbf{A} \sim \mathcal{N}^{\beta}(0, I_{d})$ and $\gamma$ be uniformly distributed on $[-\lambda, \lambda]$. For conciseness, we represent $\beta$ samples of $\gamma$ as $\Gamma \sim [-\lambda, \lambda]^{\beta}$. Using $\mathbf{A}$ and $\Gamma$, a binary sketch $h_{\mathbf{A}, \Gamma}(\mathbf{x}_{i})$ is obtained where $h_{\mathbf{A}, \Gamma}(\mathbf{x}_{i}) = \text{sign}(\mathbf{A}\mathbf{x}_i + \Gamma)$. Given sufficiently large values of $\lambda$ and $\beta$, the following holds probabilistically:

\begin{equation}\label{distancehash}
    \sup_{\mathbf{x}_{i}, \mathbf{x}_{j} \in \mathcal{X}} \Bigg| \frac{\sqrt{2 \pi} \lambda}{\beta}  d_{H}[h_{\mathbf{A}, \Gamma}(\mathbf{x}_{i}),h_{\mathbf{A}, \Gamma}(\mathbf{x}_{j})] - ||\mathbf{x}_{i}-\mathbf{x}_{j}||_{2} \Bigg| \leq \delta
\end{equation}

\noindent See Theorem 1.7 of \citet{Dirksensharp_randomHyperPlane} for details. Similar to inequality \ref{anglehash}, distances are preserved as the dimensionality of the embedding increases, assuming an appropriate choice of hyperplane shift. Note that \citet{Vershyninhyperplane_tessla} showed only an existence of such hyperplane over arbitrary sets in $\mathbb{R}^{d}$ but \citet{Dirksensharp_randomHyperPlane} showed an estimate with a bound.

\subsection{Controlling for Confounding Covariates}

As noted earlier, matching offers a way to reduce the bias introduced by confounding when estimating causal effects from observational data \citep{SUTVA_cite}. By comparing the outcomes of pairs of treated and untreated individuals that are nearly identical (i.e. matched), we can eliminate any bias arising from confounding due to any observed covariates (as well as any unobserved covariates that are sufficiently correlated with the observed covariates). The estimated individual treatment effect $\widehat{\text{ITE}}(\mathbf{x}_i)$ for individual $i$  is given by:

\begin{equation*}
        \widehat{\text{ITE}}(\mathbf{x}_i) =
    \left\{
    \begin{array}{cc}
    Y_i - Y_{Match_0(\mathbf{x}_i)} & \mathrm{if\ } T_i=1 \\
     Y_{Match_1(\mathbf{x}_i)} - Y_i & \mathrm{if\ } T_i=0 \\
    \end{array} 
    \right.
\end{equation*}

\noindent where $Match_0(\mathbf{x}_i)$ and $Match_1(\mathbf{x}_i)$ denote the indices of untreated and treated individuals that match $\mathbf{x}_i$.  The estimated unobserved outcome is denoted by $Y_{Match_T(\mathbf{x}_i)}$. If the covariates are perfectly matched, then the ignorability condition clearly holds i.e. $(Y_i^{T=0}, Y_i^{T=1}) \indep T_i | \mathbf{x}_i$ \citep{balacningrsenbaum}.

In practice, matches are seldom perfect and we must settle for nearest neighbor matching. In the context of this paper, matching is done on the $\beta$-dimensional hashed sketches of the covariate vectors, i.e., their RHPT representations.

\section{Causal Effect Estimation Using Random Hyperplane Tessellations}

We project the original $d$-dimensional covariates $\mathbf{x}_i$ into the Hamming space to get our $\beta$-dimensional hashed sketch. In our experiments, we use the \citet{Dirksensharp_randomHyperPlane} and \citet{Vershyninhyperplane_tessla} methods in order to approximately preserve both the angle and the Euclidean distance between vectors. We concatenate the results from both methods: $h_{\mathbf{A}, \Gamma}(\mathbf{x}_{i}) = \text{sign}(\mathbf{A}\mathbf{x}_i + \Gamma)$ and $h_{\mathbf{A}}(\mathbf{x}_{i}) = \text{sign}(\mathbf{A}\mathbf{x}_i)$. For brevity, from this point onward the concatenated embedding is represented by $h_{\beta}(\mathbf{x_i})$. To summarize, we first use RHPT to obtain a binary representation of the covariates, and then use nearest-neighboring matching in the binary space in order to estimate causal effects. Specifically, we use 1-Nearest Neighbor (1-NN) matching with replacement and use Hamming distance to calculate nearest neighbors.

Matching is a form of covariate adjustment, and covariate adjustment eliminates confounding under the standard assumptions stated in Section \ref{Preliminaries}. To justify matching using RHPT representations of covariate vectors,  it suffices to show that the RHPT embeddings achieve \textit{covariate balance}. We discuss this property of the RHPT embedding in the following section.

 \subsection{Approximate Balancing Scores using RHPT}\label{approx-balance-rhpt}

 A function $b(\mathbf{x}_i)$ is a balancing score if there exists a mapping from $b(\mathbf{x}_i)$ to the propensity score  $e(\mathbf{x}_i) = \mathbb{E}[T_{i}|\mathbf{x}_{i}]$; $\forall \mathbf{x}_i \in \mathcal{X}$. Let this mapping be $g(\cdot)$, where $g(b(\mathbf{x}_i)) = e(\mathbf{x}_i)$. Matching can be performed on the balancing scores rather than the raw covariates as implied by Theorems 2 and 3 from \citet{balacningrsenbaum}, i.e. $\mathbf{x}_{i} \indep T_{i}|b(\mathbf{x}_{i})$ and $Y_i(1), Y_i(0) \indep T_i|b(\mathbf{x}_i)$,  assuming the {\em Ignorability}, {\em Positivity}, {\em Non-interference} and {\em Consistency} assumptions hold on the original data.

If the embedding $b(x_i)$ is a one-to-one mapping from $\mathcal{X}$, and there is a mapping from the original covariates to the propensity scores, then there must exist a mapping from the embedding to the propensity scores $g(b(\mathbf{x}_i)) = e(\mathbf{x}_i)$. Because the function $b(\cdot)$ is one-to-one mapping, an inverse function exists, therefore, we can express the function $g(b(\mathbf{x}_i))$ as $e(b^{-1}(b(\mathbf{x}_i)))$.

 Unfortunately, for our method which uses RHPT  embeddings of covariate vectors, we cannot guarantee  one-to-one mapping using $h_{\beta}(\cdot): \mathcal{X} \rightarrow \{0,1\}^{\beta}$. Let $\mathbf{x}_{i} \in \mathcal{X}$ and $\mathbf{x}_{j} \in \mathcal{X}$, be two covariate vector that are mapped to the same binary representation using RHPT. In this case, the Hamming distance $d_{H}[h_{\beta}(\mathbf{x}_{i}),h_{\beta}(\mathbf{x}_{j})] = 0$. Substituting this term into inequalities \ref{anglehash} and \ref{distancehash}, we get:
 
 \begin{equation*}
        d_{S^{d-1}}[\mathbf{x}_i,\mathbf{x}_j] \leq \delta \quad \text{and} \quad ||\mathbf{x}_{i}-\mathbf{x}_{j}||_{2} \leq \delta 
\end{equation*}

\noindent Therefore, two covariate vectors that are within distance $\delta$ (angular and L2) from each other may be mapped to the same Hamming code. Because $h_{\beta}(\cdot)$ is not a one-to-one mapping, we cannot exactly recover these covariate vectors by ``inverting'' their representations. Particularly, if $d_{H}[h_{\beta}(\mathbf{x}_{i}),h_{\beta}(\mathbf{x}_{j})] = 0$, then the ``reconstruction'' $h^{-1}(h_{\beta}(\mathbf{x}_{i}))$ could be $\mathbf{x}_{i}$ or $\mathbf{x}_{j}$. For simplicity we say that the reconstruction maps to only one of the arbitrarily chosen points with the same binary code (i.e. either $\mathbf{x}_i$ or $\mathbf{x}_j$).

The results in section \ref{Preliminaries} state that as $\beta$ increases, $\delta \rightarrow 0$ with high probability. This means that if the Hamming codes are high-dimensional, and the Hamming codes of two points are the same, then the original points will be similar with high probability. In such a scenario, the propensity scores predicted using the reconstruction will be approximately equal if the propensity score function is assumed to be \textit{smooth}. Formally, $e(h^{-1}(h_{\beta}(\mathbf{x}_{i}))) = e(h^{-1}(h_{\beta}(\mathbf{x}_{j})) \approx e(\mathbf{x}_{i}) \approx e(\mathbf{x}_{j})$ if $e(\cdot)$ is a smooth function i.e.  $e(\mathbf{x}_{i}) \approx e(\mathbf{x}_{i} \pm \epsilon)$, where $\epsilon$ is a small perturbation close to 0. It is reasonable to assume that the propensity score function of real-world data is smooth -- this means that the probability of treatment does not change sharply across individuals with slightly different covariates vectors.

Therefore, RHPT embeddings are approximate balancing scores, given that $\beta$ is sufficiently large and $e(\cdot)$ is a smooth function. Note that there is no way of knowing the value of $\beta$ such that $h_{\beta}(\mathbf{x}_{i})$ is an approximate balancing score, since calculating the value of such a $\beta$ involves dealing with unknown constants mentioned in section \ref{Preliminaries}.

\subsection{Empirical evidence that RHPT yields approximate balancing scores}

An empirical approach used to verify if a representation is as an approximate balancing score involves fitting a function to predict propensity score given the representation, then comparing this predicted score to the true propensity score. In real-world data sets, we do not have access to true propensity scores. To address this limitation, we generate a semi-synthetic data set using the method proposed by \citet{HMNIST_ref} for the MNIST data set. For the purpose of evaluating the proposed method, we obtain the true propensity scores by simulating the data-generation process. 
%We carefully set the data-generation process parameters to ensure that there is no unobserved confounding.

To assess if our RHPT embeddings are balancing scores, we predict propensity scores using logistic regression learned on our embeddings and compare our predictions to the true propensity scores. Specifically, we calculate the mean absolute difference between the the true propensity scores and our predicted propensity scores, given by $\psi = \frac{1}{n}\sum_{i=1}^{n} |e(\mathbf{x}_{i}) - \hat{e}(h_{\beta}(\mathbf{x}_{i}))|$ where $\hat{e}(h_{\beta}(\mathbf{x}_{i}))$ is the logistic regression model trained on $h_{\beta}(\mathbf{x}_{i})$. Low values of $\psi$ indicate that our embeddings $h_{\beta}(\mathbf{x}_{i})$ are an approximate balancing scores.

\begin{figure}[t]
\includegraphics[width=8cm]{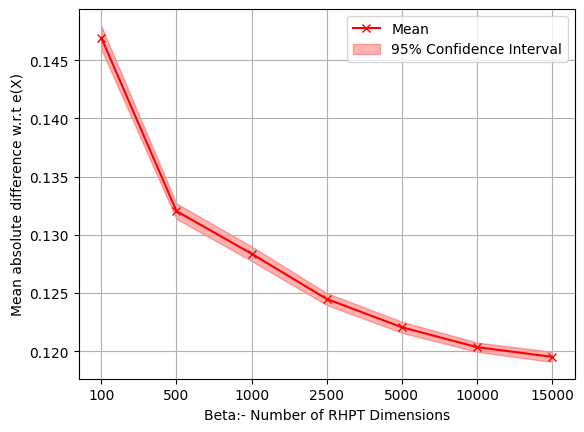}
\centering
\caption{Error of propensity score predictions $\psi$ across various RHPT embedding dimensionalities. Means and confidence intervals are computed over 100 instantiations of RHPT on a single draw of the synthetic data. Lower   $\psi$  favors ignorability.}
\label{balancing_empirical}
\end{figure}

Figure \ref{balancing_empirical} illustrates our empirical results across varying embedding dimensions $\beta$. We see that as $\beta$ increases, the mean absolute difference between predicted and true propensity scores decreases. This evaluation was performed in order to complement our theoretical arguments and provide empirical evidence that RHPT embeddings are (approximate) balancing scores. As discussed in Section \ref{approx-balance-rhpt}, causal effects can be estimated by matching on balancing scores rather than by matching on the raw covariates. The prediction error in Figure \ref{balancing_empirical} is relatively low, and decreases as the RHPT dimensionality increases. These results suggest that RHPT embeddings are approximate balancing scores and that matching using RHPT is a valid method of estimating causal effects.

\subsection{Uncertainty in causal estimates using RHPT matching}

Our proposed method makes use of a randomized algorithm to map covariates to our RHPT representation. As such, conducting experiments with different random seeds will yield different results. Specifically, different random mappings will result in different nearest-neighbor matches, thus affecting ATE and ITE predictions. Given the theoretical results from section \ref{Preliminaries}, we would expect more variation in our results for lower-dimensional RHPT embeddings. This is because distances $\delta$ are approximately preserved when the embedding dimensionality $\beta$ increases.

\begin{figure}[h]
\hfill
\subfigure[News Dataset]{\includegraphics[width=7.5cm]{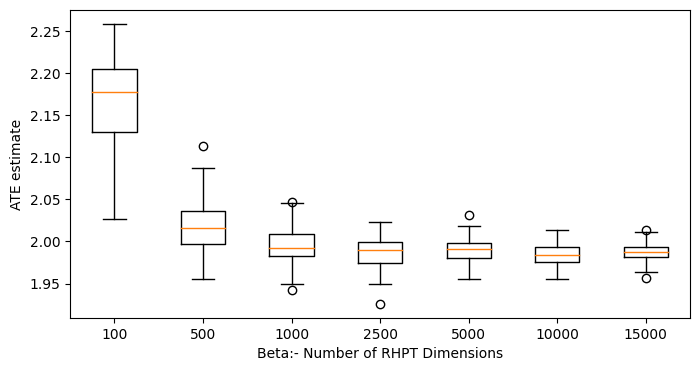}}
\hfill
\subfigure[HCMNIST Dataset]{\includegraphics[width=7.5cm]{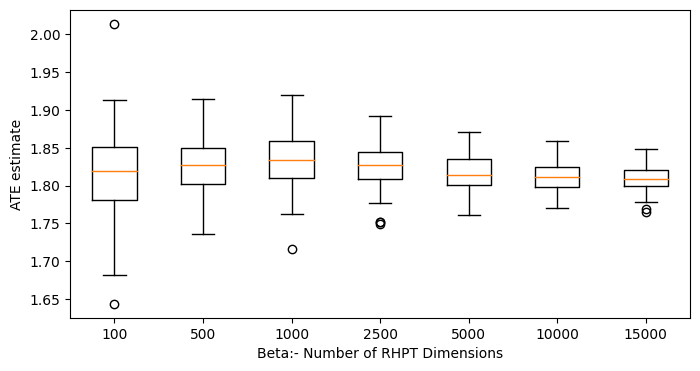}}
\hfill
\caption{Estimated ATE using RHPT on one instantiation of the synthetic dataset with various hash dimensions, each over 100 randomized runs. Higher dimensionality embeddings result in more reliable estimates.}
\label{RHPT_sensitivity}
\end{figure}

To address this, we perform experiments over 100 randomized runs of RHPT with various choice of $\beta$, with results and variability illustrated in Figure \ref{RHPT_sensitivity}. We observe that as we increase the value of $\beta$, the variability of the ATE decreases until it eventually levels off when   $\beta$ reaches $10000$. 

\section{Experiments}
We conduct experiments to compare the performance of RHPT with both classic matching techniques as well as  the state-of-the-art deep learning techniques. Baseline 1-Nearest Neighbor (1-NN) matching techniques include matching on the raw covariates $\mathbf{X}$ and matching on the PCA-transformed covariates $\mathbf{Z}$. For matching on PCA-transformed covariates we use 5 principal components following the experiments of \citet{Beyer1999WhenI}, which showed that nearest neighbor queries become unreliable when the number of principal components used exceeds 10.

Recognizing the potential limitations of PCA as a dimensionality reduction technique, we incorporate additional 1-NN matching methods using alternative dimensionality reduction techniques. Specifically, we utilize Locality Preserving Projections (LPP) proposed in \citet{LPP_He} as well as randomized projections based on the Johnson-Lindenstrauss lemma, as outlined in \citet{JL_lemma_matching}. We denote the 1-NN matching based on Locality Preserving Projections as \textbf{LPP} and the one based on randomized projections from the Johnson-Lindenstrauss lemma as \textbf{JL-Lemma}.

We also benchmark against a propensity scores which are calculated using logistic regression over the raw covariate space $\mathbf{X}$ and PCA features $\mathbf{Z}$, which we
refer to as $\hat{\text{e}}$($\mathbf{X}$) and $\hat{\text{e}}$($\mathbf{Z}$), respectively. In addition, we consider matching uniformly at random (random matching) as a simple baseline. Finally, we compare our results with those obtained using the state-of-the-art deep learning techniques: CFRNet with Maximum Mean Discrepancy (MMD) \citep{CFRnet} and DragonNet with Targeted Regularization \citep{Dragonnet}.  The codebase for our experiments is available at \url{https://github.com/Abhishek-Dalvi410/RHPT_matching}.

\subsection{Evaluation Metrics}

We compare RHPT with other methods using two different classes of performance metrics: (1) Error  of estimated causal effects (relative to their true value); and (2) Computational cost of causal effect estimation (training as well as inference).

We evaluate the error of causal effect estimates in two different settings: (1) \emph{within-sample}, where the task is to estimate the causal effect where the factual outcome is observed, and (2) \emph{out-of sample}, where the goal is to estimate the causal effect where the factual outcome is unobserved. For within-sample evaluation we estimate the true $\text{ITE}(\mathbf{x}_i)$ by direct modelling. Given an individual with observation $(\mathbf{x}_i, T_i, Y_i)$, the transductive ITE \citep{CFRnet} is given by:

\begin{equation}
        \widehat{\text{ITE}}_{in}(\mathbf{x}_i) =
    \left\{
    \begin{array}{cc}
    Y_i - g(\mathbf{x}_i, 0) & \mathrm{if\ } T_i=1 \\
      g(\mathbf{x}_i, 1) - Y_i & \mathrm{if\ } T_i=0 \\
    \end{array} 
    \right.
\end{equation}

\noindent where the function $g(\mathbf{x}_i, 1)$ and $g(\mathbf{x}_i, 0)$ are the estimate of the counterfactual for individual $i$. For CFRnet and DragonNet, $g(\mathbf{x}_i, 1)$ and $g(\mathbf{x}_i, 0)$ are neural networks, while for matching techniques they are $\hat{Y}_{NN_1(\mathbf{x}_i)}$ and $\hat{Y}_{NN_0(\mathbf{x}_i)}$, respectively.

We introduce two within-sample metrics for evaluation. The first is within-sample $\epsilon_{\text{ATE}}$, which is the mean absolute difference between the true and the predicted ATE:

$$\epsilon_{\text{ATE}} = |(\frac{1}{n} \sum_{i=0}^{n-1}\widehat{\text{ITE}}_{in}(\mathbf{x}_i)) - (\frac{1}{n} \sum_{i=0}^{n-1}\text{ITE}(\mathbf{x}_i))|$$

\noindent Note that the true effect i.e. $\text{ITE}(\mathbf{x}_i)$ considers noiseless outcomes, while the factual outcomes $Y_i$ seen in the data-set contain noise in them. We also use the RMSE error for ITE for within-sample evaluation:

$$\epsilon_{\text{ITE}} = \sqrt{\frac{1}{n}\sum_{i=0}^{n-1}(\widehat{\text{ITE}}_{in}(\mathbf{x}_i) - \text{ITE}(\mathbf{x}_i))^{2}}$$

For Out-of-sample metrics, we use we use inductive  ITE:  $\widehat{\text{ITE}}_{out}(\mathbf{x}_i) =  g(\mathbf{x}_i, 1) -  g(\mathbf{x}_i, 0)$, where both outcomes are predicted. For matching techniques, $g(\mathbf{x}_i, 1)$ and $g(\mathbf{x}_i, 0)$ are 1-NN matches from the within-sample individuals. For deep learning techniques, CFRNet and DragonNet,  $g(\mathbf{x}_i, 1)$ and $g(\mathbf{x}_i, 0)$ are predicted using neural networks. In this setting we also use out-of-sample $\epsilon_{\text{ATE}}$ -- the absolute error in the ATE estimation. The final metric we use is Precision in Estimation of Heterogeneous Effect (PEHE) from \citet{BART}, given by:

$$\epsilon_{\text{PEHE}} = \sqrt{\frac{1}{n}\sum_{i=0}^{n-1}(\widehat{\text{ITE}}_{out}(\mathbf{x}_i) - \text{ITE}(\mathbf{x}_i))^{2}}$$

\begin{table*}[ht!]

\begin{center}
\begin{tabular}{ |c||c|c|c|c|c|c| }
\hline
  & \multicolumn{2}{|c|}{Within-sample metrics} & \multicolumn{2}{|c|}{Out-of-sample metrics} & \multicolumn{2}{|c|}{Wall-Time} \\
\hline
 & $\epsilon_{\text{ATE}}$ & $\epsilon_{\text{ITE}}$  & $\epsilon_{\text{ATE}}$  & $\epsilon_{\text{PEHE}}$ & GPU & CPU\\
\hline
\hline
Random & 0.87 $\pm$ 0.08 & 4.08 $\pm$ 0.20 & 0.84 $\pm$ 0.09 & 5.72 $\pm$ 0.30 & - & 0.63 \\
\hline
X & 0.42 $\pm$ 0.06 & 3.04 $\pm$ 0.11 & 0.75 $\pm$ 0.11 & 4.22 $\pm$ 0.18 & - & 1.82\\
\hline
Z & 0.34 $\pm$ 0.03 & 2.77 $\pm$ 0.07 & 0.36 $\pm$ 0.04 & 3.86 $\pm$ 0.15 & - & 1.67 \\
\hline
$\hat{\text{e}}$(X) & 0.48 $\pm$ 0.09 & 2.88 $\pm$ 0.13 & 0.66 $\pm$ 0.13 & 5.00 $\pm$ 0.24 & - & 1.59 \\
\hline
$\hat{\text{e}}$(Z) & 0.44 $\pm$ 0.03 & 3.19 $\pm$ 0.09 & 0.47 $\pm$ 0.05 & 4.49 $\pm$ 0.18 & - & 1.66 \\
\hline
LPP & 0.49 $\pm$ 0.04 & 3.17 $\pm$ 0.10 &  0.55 $\pm$ 0.06 & 4.73 $\pm$ 0.19 & - & 21.56 \\
\hline
JL-Lemma &  0.49 $\pm$ 0.05 & 3.30 $\pm$ 0.11 & 0.54 $\pm$ 0.06 & 4.63 $\pm$ 0.19 & - & 1.05 \\
\hline
RHPT$^{\text{100}}$ & 0.45 $\pm$ 0.04 & 3.11 $\pm$ 0.12 &  0.50 $\pm$ 0.06  & 4.23 $\pm$ 0.19 & - & 1.37 \\
\hline
RHPT$^{\text{500}}$ & 0.22 $\pm$ 0.02 & 2.51 $\pm$ 0.06 &  0.27 $\pm$ 0.03  & 3.37 $\pm$ 0.12 & - & 1.93 \\
\hline
RHPT$^{\text{1k}}$ & 0.19 $\pm$ 0.02 & 2.39 $\pm$ 0.06 &  0.25 $\pm$ 0.03  & 3.18 $\pm$ 0.12 & - & 2.14 \\
\hline
RHPT$^{\text{2.5k}}$ & 0.18 $\pm$ 0.02 & 2.36 $\pm$ 0.06 &  0.25 $\pm$ 0.03  & 3.18 $\pm$ 0.12 & - & 5.16 \\
\hline
RHPT$^{\text{5k}}$ & 0.17 $\pm$ 0.02 & 2.35 $\pm$ 0.05 &  0.25 $\pm$ 0.04  & 3.15 $\pm$ 0.13 & - & 11.09 \\
\hline
RHPT$^{\text{10k}}$ & 0.16 $\pm$ 0.02 & 2.34 $\pm$ 0.05 &  0.25 $\pm$ 0.04  & 3.14 $\pm$ 0.13 & - & 16.00 \\
\hline
CFRNet$^{\text{MMD}}$ & 0.16 $\pm$ 0.01 & 2.03 $\pm$ 0.03 & 0.22 $\pm$ 0.02 & 2.63 $\pm$ 0.07 & 38.84 & $\geq$ 2000 \\
\hline
DragonNet$^{\text{T-reg}}$ & 0.13 $\pm$ 0.01 & 1.92 $\pm$ 0.04 & 0.16 $\pm$ 0.02 & 2.25 $\pm$ 0.07 & 39.26 & $\geq$ 2000 \\
\hline
\end{tabular}
\end{center}
\caption{Mean errors (and standard errors) with Wall-times (minutes) across 50 simulations of News Dataset. Rows indicate different causal effect estimation methods (randomly matching, matching on covariates X, matching on PCA-transformed covariates Z, etc).}
\label{News-table}
\end{table*}

\begin{table*}[ht!]

\begin{center}
\begin{tabular}{ |c||c|c|c|c|c|c| }
\hline
  & \multicolumn{2}{|c|}{Within-Sample Metrics} & \multicolumn{2}{|c|}{Out-of-Sample Metrics} & \multicolumn{2}{|c|}{Wall-Time} \\
\hline
 & $\epsilon_{\text{ATE}}$ & $\epsilon_{\text{ITE}}$  & $\epsilon_{\text{ATE}}$  & $\epsilon_{\text{PEHE}}$ & GPU & CPU\\
\hline
\hline
Random & 0.26 $\pm$ 0.01 & 4.04 $\pm$ 0.01 & 0.32 $\pm$ 0.02 & 4.94 $\pm$ 0.02 & - & 1.86 \\
\hline
X & 0.26 $\pm$ 0.01 & 3.59 $\pm$ 0.01 & 0.32 $\pm$ 0.02 & 3.71 $\pm$ 0.02 & - & 1.98 \\
\hline
Z & 0.24 $\pm$ 0.01 & 3.70 $\pm$ 0.01 &  0.30 $\pm$ 0.02 & 4.06 $\pm$ 0.02 & - & 1.92 \\
\hline
$\hat{\text{e}}$(X) & 0.25 $\pm$ 0.02 & 4.03 $\pm$ 0.01 & 0.39 $\pm$ 0.03 & 4.97 $\pm$ 0.02 & - & 8.38 \\
\hline
$\hat{\text{e}}$(Z) & 0.23 $\pm$ 0.01 & 3.98 $\pm$ 0.01 & 0.31 $\pm$ 0.02 & 4.64 $\pm$ 0.02 & - & 1.97 \\
\hline
LPP & 0.27 $\pm$ 0.01 & 3.91 $\pm$ 0.01 &  0.34 $\pm$ 0.03 & 4.85 $\pm$ 0.03 & - & 5.68 \\
\hline
JL-Lemma &  0.26 $\pm$ 0.01 & 3.81 $\pm$ 0.01 & 0.36 $\pm$ 0.02 & 4.29 $\pm$ 0.02 & - & 2.13 \\
\hline
RHPT$^{\text{100}}$ & 0.23 $\pm$ 0.01 & 3.68 $\pm$ 0.01 & 0.29 $\pm$ 0.02 & 3.95 $\pm$ 0.02 & - & 1.96 \\
\hline
RHPT$^{\text{500}}$ & 0.21 $\pm$ 0.01 & 3.63 $\pm$ 0.01 & 0.24 $\pm$ 0.02 & 3.80 $\pm$ 0.02 & - & 2.55 \\
\hline
RHPT$^{\text{1k}}$ & 0.20 $\pm$ 0.01 & 3.61 $\pm$ 0.01 & 0.24 $\pm$ 0.02 & 3.77 $\pm$ 0.02 & - & 3.08 \\
\hline
RHPT$^{\text{2.5k}}$ & 0.20 $\pm$ 0.01 & 3.61 $\pm$ 0.01 & 0.23 $\pm$ 0.02 & 3.75 $\pm$ 0.02 & - & 3.59 \\
\hline
RHPT$^{\text{5k}}$ & 0.20 $\pm$ 0.01 & 3.60 $\pm$ 0.01 & 0.23 $\pm$ 0.02 & 3.75 $\pm$ 0.02 & - & 6.18 \\
\hline
RHPT$^{\text{10k}}$ & 0.19 $\pm$ 0.01 & 3.58 $\pm$ 0.01 & 0.23 $\pm$ 0.02 & 3.73 $\pm$ 0.02 & - & 8.30 \\
\hline
CFRNet$^{\text{MMD}}$  &  0.21 $\pm$ 0.01 & 3.31 $\pm$ 0.01 & 0.25 $\pm$ 0.02 & 3.09 $\pm$ 0.02 & 56.34 & 181.43 \\
\hline
DragonNet$^{\text{T-reg}}$ & 0.23 $\pm$ 0.01 & 2.84 $\pm$ 0.00 & 0.24 $\pm$ 0.02 & 2.16 $\pm$ 0.01 & 21.01 & 59.41 \\

\hline
\end{tabular}
\end{center}
\caption{Mean errors (and standard errors) with wall-times in minutes across 100 simulations of HCMNIST Dataset. Rows indicate different causal effect estimation methods (randomly matching, matching on covariates X, matching on PCA-transformed covariates Z, etc).}
\label{HCMNIST-table}
\end{table*}

We compare the computational cost of different methods using run-time (training and inference) for each method. The deep learning methods are run on two types of machines: (1) using an NVIDIA A-40 GPU and (2) using the free version of Google Colab in CPU mode (with no GPU). Matching techniques are not evaluated using a GPU since matching does not require computationally intensive training. Note that for the free version of Google Colab in CPU mode, there is a limit on the of execution time. If that limit is exceeded during execution, we report a run-time $\geq 2000$ minutes.

\subsection{Data sets}

Data sets used for evaluation are semi-synthetic, meaning that the covariates come from real-world data while the treatments and outcomes were generated through a user-specified data generating process. For each data set, we report the average and standard error of the within-sample and out-of-sample errors over different instantiations of the data generating process. The wall-time is reported as the total time required to run each method on of all samples from the data generating process. For all data sets, 10\% of the samples are used for out-of-sample evaluation. For deep learning models, 20\% of in-sample data points are used as validation set to avoid over-fitting.

\subsubsection{News}

\citet{CFRnet} used a simulated data set to replicate how opinions and perceptions toward news articles could differ based on whether they accessed the content on a desktop (T=0) or a mobile device (T=1). In each draw of the data generating process, 5000 news articles are randomly sampled from the NY Times corpus \citep{BAGwords}. The covariates $\mathbf{x}_i$ of each news article is a Bag of Words with dimensionality 3477. Using the covariates, treatment assignments (desktop (T=0) or mobile device (T=1)) and outcomes (opinions) for each news articles are generated. We use 50 draws of the Data Generating Process provided by \citet{CFRnet}.

\subsubsection{HC-MNIST} \citet{HMNIST_ref} introduced a semi-synthetic data set that based on the MNIST dataset \citet{MNIST_ref}. The covariates are the 28 $\times$ 28 pixel values of the images with total dimensionality of 784. The representation of the images are used to generate binary treatment and continuous outcomes. The data generating process contains $\Gamma \geq 1$ factor which controls the degree of hidden confounding with $\Gamma = 1$ representing no unobserved confounding.  For each instance/draw out of the 100 draws of the simulation, the treatment and outcomes are generated according to \citet{HMNIST_ref} method and then 3000 images are randomly selected. Only 3000 samples/images are selected out of the complete data set (42000 images) to introduce sparsity in the data space so that we can emulate the curse of dimensionality. We set $\Gamma = 1.1$ to mimic a real world scenario. (Full unconfoundedness is not plausible in the real world).

In addition to News and HC-MNIST dataset, we also perform experiment on 100 draws of Infant Health and Development Program (IHDP) dataset by \citet{BART} which includes covariates from a randomized controlled trial devised to assess how cognitive test scores in premature infants are impacted by visits from specialized doctors. IHDP data set is a low dimensional dataset with 25 covariates of which 16 are binary and 9 are continuous. While this dataset falls outside the primary scope of our algorithm due to its low dimensionality, we have included its results in the appendix to demonstrate that our method remains functional even for low-dimensional data.

\subsection{Hyperparameters for Deep learning models}

The hyperparameters for deep learning models are presented in Table \ref{DL_HP_table}. The hyperparameters are found using the method described in \citet{TARNET}. The loss functions of these deep learning models consist of multiple terms. All these terms are weighted equally i.e. weights/imbalance parameter are 1. In all the models, the initial two layers don't have L2 regularization; every other layer has L2 regularization weighted by 0.01. One significant observation is that the hyper-parameters for early stopping and learning rate were more important as compared to the neural network architecture for achieving strong performance. For training CFRNet, we use Adam optimizer \citep{Kingma2014AdamAM} and for DragonNet we use Nesterov accelerated gradient descent (NAG) \citep{Nesterov1983AMF} with momentum.

\begin{table*}[ht!]
\begin{center}
\begin{tabular}{ |c||c|c|c|c|c| }
\hline
 &  &   & Early & Starting & LR  \\
 Dataset & Model & Layers  & Stopping & Learning & Reduction  \\
 &  &   & Patience & Rate (LR) & Patience  \\
\hline

\multirow{2}{*}{News} & CFRNET  & [400,400,[200,200]] & 40 & 2.5e-3  & 10 \\
 \cline{2-6}
    & Dragonnet   & [200,200,[100,100]] & 40 & 1e-7 & 5 \\
 \hline

 \multirow{2}{*}{HCMNIST} & CFRNET  & [512,256,[256,128]] & 30 &  2.5e-3 & 10 \\
 \cline{2-6}
    & Dragonnet   & [512,256,128,[64,32]] & 20 & 1e-7 & 5 \\
 \hline

 \hline

\end{tabular}
\end{center}
\caption{Hyperparameters for the deep learning models}
\label{DL_HP_table}
\end{table*}

\subsection{Results}

For the News dataset, we observe that RHPT matching significantly outperforms traditional matching techniques on all metrics. Additionally, it is noteworthy that the Within-sample $\epsilon_{\text{ATE}}$ error for both RHPT matching and deep learning models is almost the same, while the $\epsilon_{\text{ITE}}$ error and the Out-of-sample $\epsilon_{\text{ATE}}$ error is slightly worse for RHPT, but still within the same ballpark as the deep learning models. There is a significant difference between RHPT and the deep learning models for the $\epsilon_{\text{PEHE}}$ error. RHPT matching outperforms CFRNet and DragonNet in terms of the wall-time needed for estimating the causal effect. Due to high dimensionality a large sample size of  the News dataset, it takes a significant amount of time to execute on a CPU when using deep learning models. Even when a GPU is used, the wall time for RHPT matching is less than half of the GPU wall time of the deep learning models.

As seen from Table \ref{HCMNIST-table}, HCMNIST dataset appears to be more challenging for causal effect estimation than the News dataset, as the improvement from traditional matching techniques to RHPT and deep learning models is not as pronounced as it is for the News dataset. The In-sample $\epsilon_{\text{ATE}}$ error for all methods is nearly the same, with RHPT performing the best. The out-of-sample $\epsilon_{\text{ATE}}$ error is significantly better for RHPT matching when compared to traditional matching techniques and on par with the deep learning models. The $\epsilon_{\text{PEHE}}$ error is significantly better for RHPT matching than traditional matching techniques, but deep learning methods outperform any of the matching techniques.

RHPT outperforms traditional matching techniques in terms of both in-sample and out-of-sample metrics when the dimensionality of the RHPT representation is sufficiently large ($\geq 10k$) with only modest increase in computational cost. We see that our matching technique significantly and consistently outperforms all traditional matching techniques when it comes to the $\epsilon_{\text{ATE}}$, $\epsilon_{\text{ITE}}$ and $\epsilon_{\text{PEHE}}$   and is competitive with CFRNet and DragonNet. Deep learning methods slightly outperform RHPT in terms of out-of-sample $\epsilon_{\text{PEHE}}$ error; and RHPT is competitive with deep learning methods in terms of within-sample error. In both cases,  RHPT matching incurs computational cost that is an order or two magnitude lower than deep learning methods.

\section{Conclusion and Discussion}
We have introduced a simple, fast, yet effective approach to estimating causal effects from high-dimensional observational data. The proposed method makes use of  Random Hyperplane Tessellations (RHPT) to map covariates into a high-dimensional, binary space based on the relationship between covariate values and randomly chosen hyperplanes. We show that RHPT ensures approximate covariate balance, a theoretical requirement for accurate causal effect estimation from observational data.  We report results of extensive experiments showing that matching using RHPT outperforms traditional matching techniques and is competitive with state-of-the-art deep learning methods for causal effect estimation. Matching using RHPT also provides several important benefits not offered by deep learning methods. Namely, RHPT avoids the need for computationally expensive training, does not require large amounts of data, and requires little hyperparameter tuning.

A promising direction for future research include extensions of the proposed method to causal effect estimation in settings where the observations are not independent and identically distributed, e.g., when individuals are linked by social or other ties. Another interesting direction is deeper theoretical analyses of the conditions under which RHPT or similar methods can be expected to yield accurate causal effect estimates. Finally, causal inference in the online learning and continual learning settings remains relatively unexplored. Matching-based approaches such as RHPT are well-suited for this setting, as they do not require iterative training and can be easily updated as new data becomes available.

\section*{Acknowledgments}
This work was funded in part by grants from the National Science Foundation (2226025, 2041759), the National Center for Advancing Translational Sciences, and the National Institutes of Health (UL1 TR002014). Deep Learning GPU Computations for this research were performed on the Pennsylvania State University’s Institute for Computational and Data Sciences’ Roar supercomputer.

\bibliographystyle{plainnat}
\bibliography{references}  %%% Uncomment this line and comment out the ``thebibliography'' section below to use the external .bib file (using bibtex) .





% \end{thebibliography}

\appendix

\section{Additional experimental results}
Results for the IHDP dataset are provided in Table \ref{IHDP results}. RHPT matching provides similar performance to matching on the raw covariates. Since the IHDP dataset is relatively low-dimensional with only 25 features, the curse-of-dimensionality does not hinder matching performance, therefore RHPT matching does not provide much added benefit.

\begin{table*}[ht!]
\begin{center}
\begin{tabular}{ |c||c|c|c|c|c|c| }
\hline
  & \multicolumn{2}{|c|}{Within-Sample Metrics} & \multicolumn{2}{|c|}{Out-of-Sample Metrics} & \multicolumn{2}{|c|}{Wall-Time} \\
\hline
 & $\epsilon_{\text{ATE}}$ & $\epsilon_{\text{ITE}}$  & $\epsilon_{\text{ATE}}$  & $\epsilon_{\text{PEHE}}$ & GPU & CPU\\
\hline
\hline
Random & 0.30 $\pm$ 0.04 & 4.23 $\pm$ 0.50 & 0.94 $\pm$ 0.18 & 8.47 $\pm$ 1.16 & - & 0.07 \\
\hline
X & 0.14 $\pm$ 0.02 & 2.64 $\pm$ 0.25 & 0.67 $\pm$ 0.13 & 4.85 $\pm$ 0.63 & - & 0.12 \\
\hline
Z & 0.17 $\pm$ 0.02 & 3.0 $\pm$ 0.30 & 0.70 $\pm$ 0.16 & 5.63 $\pm$ 0.74 & - & 0.12 \\
\hline
e(X) & 0.22 $\pm$ 0.02 & 4.16 $\pm$ 0.5 & 0.79 $\pm$ 0.14 & 8.23 $\pm$ 1.12 & - & 0.13 \\
\hline
e(Z) & 0.20 $\pm$ 0.02 & 4.02 $\pm$ 0.47 & 0.8 $\pm$ 0.12 & 7.93 $\pm$ 1.08 & - & 0.20 \\
\hline
RHPT$^{\text{100}}$ & 0.14 $\pm$ 0.02 & 2.98 $\pm$ 0.31 & 0.57 $\pm$ 0.10 & 5.75 $\pm$ 0.77 & - & 0.19 \\
\hline
RHPT$^{\text{500}}$ & 0.14 $\pm$ 0.02 & 2.84 $\pm$ 0.28 & 0.55 $\pm$ 0.08 & 5.20 $\pm$ 0.71& - & 0.19 \\
\hline
RHPT$^{\text{1k}}$ & 0.15 $\pm$ 0.02 & 2.79 $\pm$ 0.25 & 0.53 $\pm$ 0.07 & 5.11 $\pm$ 0.69 & - & 0.19 \\
\hline
RHPT$^{\text{2.5k}}$ & 0.14 $\pm$ 0.01 & 2.76 $\pm$ 0.27 & 0.53 $\pm$ 0.07 & 4.98 $\pm$ 0.63 & - & 0.32 \\
\hline
RHPT$^{\text{5k}}$ & 0.14 $\pm$ 0.01 & 2.76 $\pm$ 0.28 & 0.49 $\pm$ 0.07 & 4.91 $\pm$ 0.60 & - & 0.45 \\
\hline
RHPT$^{\text{10k}}$ & 0.14 $\pm$ 0.02 & 2.75 $\pm$ 0.27 & 0.48 $\pm$ 0.10 & 4.92 $\pm$ 0.65 & - & 0.72 \\
\hline
CFRNet$^{\text{MMD}}$ & 0.10 $\pm$ 0.01 & 1.21 $\pm$ 0.04 & 0.19 $\pm$ 0.03 & 1.15 $\pm$ 0.17 & 19.58 & 117.76 \\
\hline
DragonNet$^{\text{T-reg}}$ & 0.11 $\pm$ 0.00 & 1.17 $\pm$ 0.02 & 0.16 $\pm$ 0.02 & 0.96 $\pm$ 0.12 & 47.25 & 513.32 \\
\hline

\hline
\end{tabular}
\end{center}
\caption{Mean errors (and standard errors) with Wall-times in minutes across 100 simulations of IHDP Dataset. Rows indicate different causal effect estimation methods (randomly matching, matching on covariates X, matching on PCA-transformed covariates Z, etc).}
\label{IHDP results}
\end{table*}

\end{document}